\documentclass[conference]{IEEEtran}
\IEEEoverridecommandlockouts
\usepackage{cite}
\usepackage{amsmath,amssymb,amsfonts}
\usepackage{algorithmic}
\usepackage{graphicx}
\usepackage{textcomp}
\usepackage{diagbox}
\usepackage{booktabs}

\usepackage{algorithm}
\usepackage{multirow}
\usepackage{enumitem}
\usepackage{bbding}
\usepackage{amsmath}
\usepackage{amssymb}

\usepackage{soul}
\usepackage{color, xcolor}
\usepackage{newfloat}
\usepackage{listings}

\usepackage{hyperref}
\hypersetup{hidelinks}
\hypersetup{
	colorlinks=true,
	linkcolor=black
}
\usepackage{fontawesome}

\def\BibTeX{{\rm B\kern-.05em{\sc i\kern-.025em b}\kern-.08em
    T\kern-.1667em\lower.7ex\hbox{E}\kern-.125emX}}
\begin{document}

\title{Multi-scale Bottleneck Transformer for \\ Weakly Supervised Multimodal Violence Detection}
\author{\IEEEauthorblockN{Shengyang Sun \quad\quad\quad\quad\quad Xiaojin Gong\textsuperscript{\Envelope}\thanks{\Envelope~Corresponding author.}}
	\IEEEauthorblockA{College of Information Science \& Electronic Engineering \\
		Zhejiang University, Hangzhou, Zhejiang, China\\
		 shengyang.sunshy@gmail.com, gongxj@zju.edu.cn}
}

\maketitle

\begin{abstract}
	Weakly supervised multimodal violence detection aims to learn a violence detection model by leveraging multiple modalities such as RGB, optical flow, and audio, while only video-level annotations are available. In the pursuit of effective multimodal violence detection (MVD), information redundancy, modality imbalance, and modality asynchrony are identified as three key challenges. In this work, we propose a new weakly supervised MVD method that explicitly addresses these challenges. Specifically, we introduce a multi-scale bottleneck transformer (MSBT) based fusion module that employs a reduced number of bottleneck tokens to gradually condense information and fuse each pair of modalities and utilizes a bottleneck token-based weighting scheme to highlight more important fused features. Furthermore, we propose a temporal consistency contrast loss to semantically align pairwise fused features. Experiments on the largest-scale XD-Violence dataset demonstrate that the proposed method achieves state-of-the-art performance. Code is available at \faGithubAlt~\href{https://github.com/shengyangsun/MSBT}{https://github.com/shengyangsun/MSBT}.
\end{abstract}

\section{Introduction}
Video violence detection aims to identify violent events in videos.
Recently, with the availability of multimodal data, \textit{e.g.} the audio-visual dataset XD-Violence~\cite{wu2020not}, there has been increasing attention drawn to multimodal violence detection (MVD)~\cite{yu2022modality,wu2022weakly}.
In the pursuit of effective multimodal violence detection, we have identified three key challenges that need to be addressed: 1) \textit{Information redundancy}: Each modality contains redundant information and these redundancies may introduce undesirable semantic bias~\cite{nagrani2021attention}. 2) \textit{Modality imbalance}: The information content within one modality may significantly outweigh that of another modality~\cite{du2021improving}. Treating each modality equally may lead to a degradation in detection performance. 3) \textit{Modality asynchrony}: Temporal inconsistencies may exist among different modalities even if the signals are synchronized. For example, in a typical ``abuse'' event, the abuser hits the victim first, followed by a scream~\cite{yu2022modality}.
Existing MVD methods mainly focus on utilizing audio-visual modalities and have attempted to address one or more of the aforementioned challenges. For instance, in the work of~\cite{zhang2023exploiting}, various attention mechanisms were employed to attend to the most relevant information, reducing background distractions and information redundancy.  

\begin{figure}[t]
	\centering
	\includegraphics[width=0.94\linewidth]{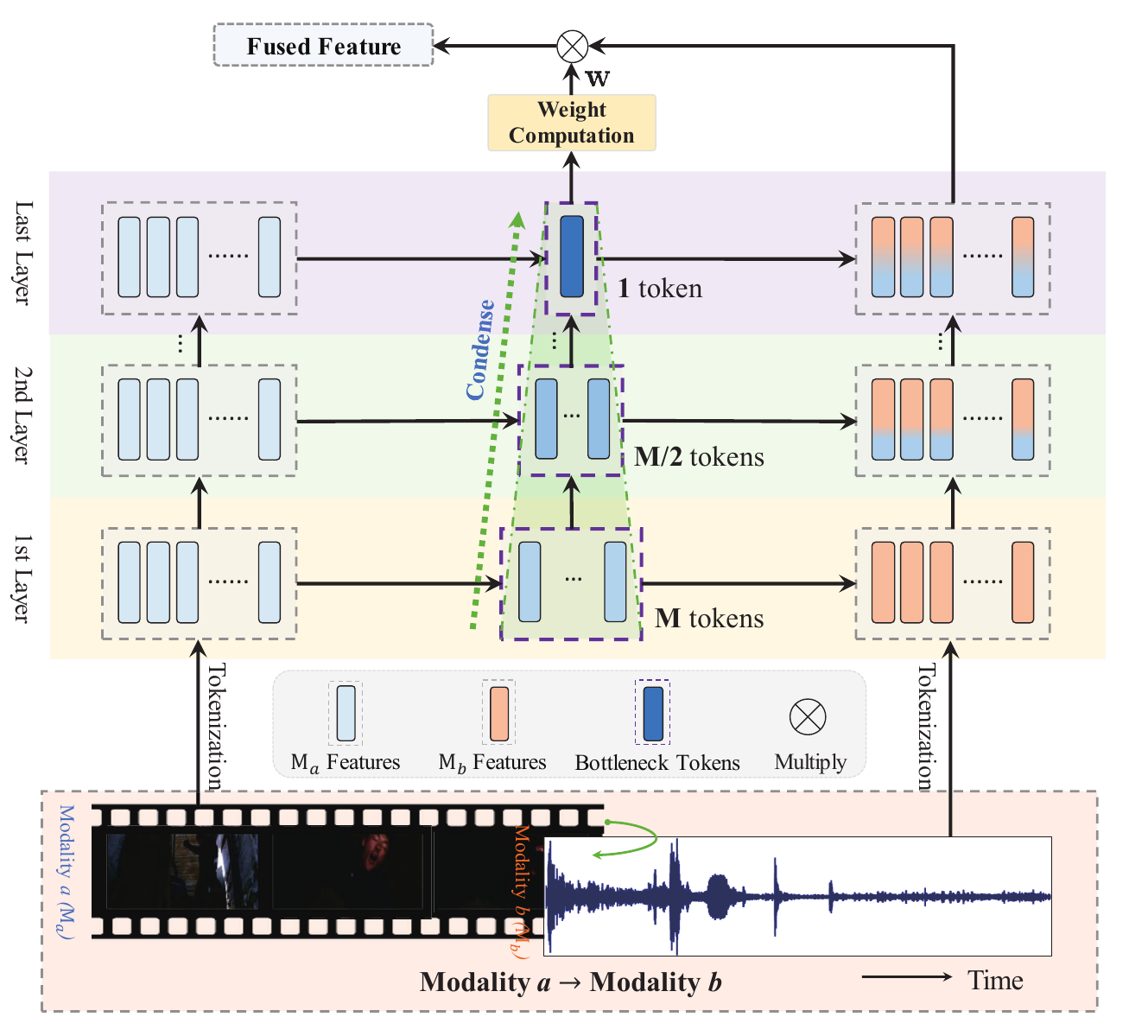}
	\caption{An illustration of our multimodal fusion module. It consists of a multi-scale bottleneck transformer and a bottleneck token-based weighting scheme. When a pair of modalities are input, the bottleneck tokens first condense the information of modality $\mathtt{a}$ and then transmit it to modality $\mathtt{b}$ at each layer. Moreover, the bottleneck tokens condensed at a layer are passed to the subsequent layer for further condensation. The token at the final layer can be used to measure the quantity of information transmitted and is therefore leveraged to weight the fused feature. Best viewed in color.}
\label{fig:index}
\vspace{-14pt}
\end{figure}

To address the first two challenges, we propose a multimodal fusion module, as illustrated in Fig.~\ref{fig:index}. This module comprises a multi-scale bottleneck transformer (MSBT) and a bottleneck token-based weighting scheme to fuse each pair of modalities. Inspired by bottleneck transformers~\cite{nagrani2021attention,Yan2022Transformer} that have been successfully applied to audio-visual classification, our MSBT is designed to employ a small set of bottleneck tokens to transmit condensed information from one modality to the other. However, unlike~\cite{nagrani2021attention,Yan2022Transformer}, we introduce a gradual condensation scheme by reducing the bottleneck tokens at consecutive layers, gradually reducing information redundancy. Furthermore, realizing that the bottleneck tokens learned at the final layer also indicate the information content transmitted between modalities, we design the bottleneck token-based weighting scheme to weight pairwise fused features, avoiding treating modalities equally. 

To address the third challenge, we propose a temporal consistency contrast (TCC) loss, which is designed based on the following observations. On one hand, the majority of audio-visual data are intrinsically synchronized, although a portion of them may be asynchronous~\cite{yu2022modality}. On the other hand, even if modality asynchrony exists, our transformer-based fusion scheme implicitly aligns each pair of modalities in the temporal dimension. Therefore, the TCC loss is designed to further semantically align all pairwise fused features at the same moment of videos, effectively addressing modality asynchrony.
We build a fully transformer-based network architecture that integrates the MSBT-based fusion module, together with the MIL loss and TCC loss, to learn a multimodal violence detection model. Our method distinguishes itself in the following aspects.
\begin{itemize}
	\item We propose a multi-scale bottleneck transformer (MSBT)-based fusion module. It leverages a reduced number of bottleneck tokens to transmit gradually condensed information from one modality to another and a bottleneck token-based weighting scheme to weight the fused features, effectively addressing the information redundancy and modality imbalance problems.
	\item We propose a temporal consistency contrast (TCC) loss to semantically align pairwise fused features at the same moment of videos, effectively handling the modality asynchrony issue.
	\item Experimental results on XD-Violence demonstrate that the proposed method achieves state-of-the-art performance, particularly when RGB, flow, and audio modalities are all considered. Furthermore, our method is extendable to any number of input modalities, potentially allowing for the inclusion of more modalities such as depth or infrared video streams.
\end{itemize}

\section{Related Works}
\subsection{Weakly Supervised Multimodal Violence Detection}
Violence detection is closely related to anomaly detection. Recently, the availability of multimodal data, particularly the release of the large-scale multimodal video dataset~\cite{wu2020not}, has sparked increasing interest in multimodal violence/anomaly detection~\cite{yu2022modality,wu2022weakly,zhang2023exploiting,xiao2022optical}. Numerous efforts have been dedicated to the fusion of audio and visual modalities, resulting in the development of various fusion methods, such as concatenation~\cite{wu2020not} and cross-attention~\cite{yu2022modality} based techniques. While the attention mechanisms employed in these works implicitly address information redundancy, there are few works explicitly designed to tackle the challenges of information redundancy, modality imbalance, and modality asynchrony, except for~\cite{yu2022modality} that handled one or two of these challenges. 

\begin{figure*}[t]
	\centering
	\includegraphics[width=0.96\textwidth]{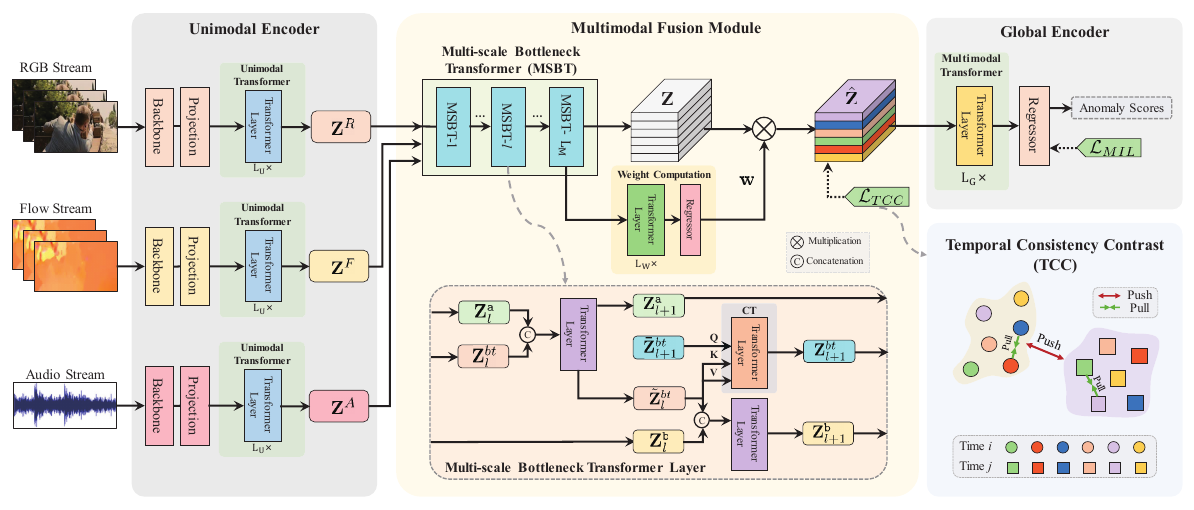}
	\caption{An overview of the proposed framework. It includes three unimodal encoders, a multimodal fusion module, and a global encoder for multimodal feature generation. Each unimodal encoder consists of a modality-specific feature extraction backbone and a linear projection layer for tokenization and a modality-shared transformer for context aggregation within one modality. The fusion module contains a multi-scale bottleneck transformer (MSBT) to fuse any pair of modalities and a sub-module to weight concatenated fused features. 
	The global encoder, implemented by a transformer, aggregates context over all modalities. Finally, the produced multimodal features are fed into a regressor to predict anomaly scores. The entire network is learned with a multiple instance learning (MIL) loss $\mathcal{L}_{MIL}$, together with a temporal consistency contrast (TCC) loss $\mathcal{L}_{TCC}$. Best viewed in color.
	}
	\label{fig:framework}
	\vspace{-14pt}
\end{figure*}

\subsection{Multimodal Transformers}
Due to the intrinsic traits of self-attention and its variants, transformers are capable of working in a modality-agnostic manner and are scalable to multiple modalities~\cite{song2022multimodal,Shvetsova2022Transformer}. However, although the representations of heterogeneous modalities are unified within transformers, challenges still persist in achieving effective fusion due to information redundancy and modality imbalance. To address the information redundancy issue, Nagrani~\textit{et al.}~\cite{nagrani2021attention} proposed a multimodal bottleneck transformer (MBT) that leverages a small set of bottleneck tokens for modality interaction. Inspired by it, we propose a multi-scale bottleneck transformer (MSBT). Unlike MBT which employs a fixed number of bottleneck tokens, our MSBT utilizes a reduced number of tokens to achieve better information condensation. Additionally, considering the modality imbalance issue, we propose to exploit the learned tokens, which indicate the transmitted information content, to weight the fused features.

\subsection{Multimodal Contrastive Learning}
Contrastive learning has been successfully applied to achieve semantic alignment of different modalities in various tasks~\cite{Shvetsova2022Transformer,yu2022modality,sun2023hierarchical}. For instance, Shvetsova~\textit{et al.}~\cite{Shvetsova2022Transformer} designed a combinatorial contrastive loss that considers all combinations of input modalities, including both single modalities and pairs of modalities, for multimodal video retrieval. Yu~\textit{et al.}~\cite{yu2022modality} proposed a modality-aware contrastive instance learning that performs cross-modal contrast for audio-visual violence detection. In this work, we design a TCC loss that treats fused features at the same moment as positive samples and features at different times as negative ones. In contrast to these works, our loss only considers the features fused from each pair of modalities while leaving out single modality features due to possible modality asynchrony. 

\section{Methodology}
This work aims to address weakly supervised violence detection when a set of untrimmed videos, associated with RGB, optical flow, and audio streams, are given. To this end, 
we propose a multimodal violence detection framework presented in Fig.~\ref{fig:framework}. As a common practice~\cite{sultani2018real}, each video is divided into $T$ non-overlapping snippets. The RGB, flow, or audio snippets go through an unimodal encoder composed of a pre-trained feature extraction backbone and a linear projection layer 
for tokenization, by which modality-specific features are generated. Then, the modality-specific features are fed into a fusion module that employs the MSBT to fuse features pairwise for all modality pairs and further concatenates and weights the fused features. The fused features finally pass through a global encoder to aggregate context and then input to a regressor to predict anomaly scores.

\subsection{Unimodal Encoder}
We first construct three unimodal encoders to generate modality-specific features for input snippets. An input snippet of one modality is tokenized by a pre-trained feature extraction backbone (\textit{e.g.}, the I3D model~\cite{carreira2017quo} for RGB or optical flow and the VGGish model~\cite{gemmeke2017audio} for audio), followed by a linear projection layer that outputs an embedded feature vector $\mathbf{z} \in\mathbb{R}^{1\times D_E}$, where ${D_E}$ is the feature dimension unified for all modalities. Then, we employ a vanilla transformer~\cite{vaswani2017attention} to aggregate context information within each modality. That is, the embedded feature is fed into an unimodal transformer with $L_U$ layers, in which each layer consists of Multi-head Self-Attention (MSA) and FeedForward Network (FFN) blocks. The computation of the $l$-th transformer layer, $\mathbf{z}_{l+1} = \text{Transformer}(\mathbf{z}_{l})$, is defined by
\begin{equation} \label{eq:MSA_FFN}
	\begin{aligned}
		\hat{\mathbf{z}}_{l} &= \text{MSA}(\text{LN}(\mathbf{z}_{l})) + \mathbf{z}_{l}, \\
		\mathbf{z}_{l+1} &= \text{FFN}(\text{LN}(\hat{\mathbf{z}}_{l})) + \hat{\mathbf{z}}_{l},
	\end{aligned}
\end{equation}
in which $\text{LN}(\cdot)$ is layer normalization. MSA and FFN are implemented following previous work~\cite{vaswani2017attention}. Note that, in the unimodal encoders, the feature extractor and linear projection layer are modality-specific, while the unimodal transformer is shared by all modalities.

\subsection{Multimodal Fusion}
Let us denote the modality-specific features of a video generated by the unimodal encoders as $\mathbf{Z}^R$, $\mathbf{Z}^F$, and $\mathbf{Z}^A$, respectively, for RGB, flow, and audio modalities. Here, $\mathbf{Z}^{\mathtt{a}} = [\mathbf{z}^{\mathtt{a}}_{{L_{U+1}}(1)}, ..., \mathbf{z}^{\mathtt{a}}_{{L_{U+1}}(T)}]\in \mathbb{R}^{T\times D_E}$, in which $\mathtt{a}\in \mathcal{M}=\{R, F, A\}$ denotes one modality, $T$ is the number of snippets, and $\mathbf{z}^{\mathtt{a}}_{{L_{U+1}}}$ is the output of the unimodal transformer. We then perform multimodal fusion via the MSBT and a bottleneck token-based weighting scheme as follows.

\noindent\textbf{Multi-scale Bottleneck Transformer.}
Bottleneck transformers~\cite{nagrani2021attention,Yan2022Transformer} have demonstrated their efficiency and effectiveness in multimodal fusion, in which a small number of bottleneck tokens are exploited to condense information of a modality, greatly reducing information redundancy. Inspired by these works, we propose a multi-scale bottleneck transformer (MSBT) with $L_M$ layers to fuse features of each two modalities. The multi-scale strategy exploits a reducing number of bottleneck tokens at different layers to condense information gradually, which is distinguished from the use of a fixed number of tokens for all layers like~\cite{nagrani2021attention,Yan2022Transformer}, achieving more effective fusion performance. More specifically, given modality-specific features $\mathbf{Z}^{\mathtt{a}}$, $\mathbf{Z}^{\mathtt{b}}$ of two different modalities $\mathtt{a}, \mathtt{b} \in \mathcal{M}$, the MSBT fuses the information of modality $\mathtt{a}$ to modality $\mathtt{b}$ at layer $l$ by
\begin{equation} \label{eq:condense_1}
	[\mathbf{Z}^{\mathtt{a}}_{l+1}, \tilde{\mathbf{Z}}^{bt}_l] = \text{Transformer}([\mathbf{Z}^{\mathtt{a}}_{l}||\mathbf{Z}^{bt}_l]),
\end{equation}
\begin{equation} \label{eq:condense_2}
	[\mathbf{Z}^{\mathtt{b}}_{l+1}, \hat{\mathbf{Z}}^{bt}_{l}] = \text{Transformer}([\mathbf{Z}^{\mathtt{b}}_{l}|| \tilde{\mathbf{Z}}^{bt}_l]), 
\end{equation}
in which $[\cdot||\cdot]$ denotes the concatenation operation. $\mathbf{Z}^{bt}_l\in\mathbb{R}^{N^{bt}_{l}\times D_E}$ ($N^{bt}_{l} \ll T$) represents the bottleneck tokens input to layer $l$, which is learned at the previous layer ($\mathbf{Z}^{bt}_1$ is randomly initialized), carrying the previously condensed information of modality $\mathtt{a}$. At the current layer, these tokens act as intermediates for transmitting information from modality $\mathtt{a}$ to modality $\mathtt{b}$. To this end, they are first refined by re-aggregating information from modality $\mathtt{a}$ via Eq.~\ref{eq:condense_1} and then transmit the information to modality $\mathtt{b}$ by Eq.~\ref{eq:condense_2}, through which the fusion result $\mathbf{Z}^{\mathtt{b}}_{l+1}$ is obtained.

On the other hand, the bottleneck tokens in our MSBT also play a role in passing the condensed information of modality $\mathtt{a}$ from the current layer to the next layer. To achieve this, we additionally initialize a new set of bottleneck tokens $\bar{\mathbf{Z}}^{bt}_{l+1}\in\mathbb{R}^{N^{bt}_{l+1}\times D_E}$ at layer $l$. The number of tokens is reduced by half, \textit{i.e.}, $N^{bt}_{l+1} = \lfloor N^{bt}_{l}/2\rfloor$, to store more condensed information. The condensed information of layer $l$ is passed to layer $l+1$ by a cross-attention based transformer:
\begin{equation} \label{eq:cross_transformer}
	\mathbf{Z}^{bt}_{l+1} = \text{Cross-Transformer}(\bar{\mathbf{Z}}^{bt}_{l+1}, \tilde{\mathbf{Z}}^{bt}_l),
\end{equation}
where $\text{Attention}(\mathbf{x}\mathbf{W}^Q, \mathbf{y}\mathbf{W}^K, \mathbf{y}\mathbf{W}^V)$ is computed in the $\text{Cross-Transformer}(\mathbf{x},\mathbf{y})$. The pairwise fusion procedure is illustrated in Fig.~\ref{fig:index} and Fig.~\ref{fig:framework}. When fusing modality-specific feature $\mathbf{Z}^{\mathtt{a}}$ to $\mathbf{Z}^{\mathtt{b}}$, we take $\mathbf{Z}^{\mathtt{b}}_{L_M+1}$ as the fusion result and denote it as $\mathbf{Z}^{\mathtt{a}\mathtt{b}}$. This essentially transmits condensed information of modality $\mathtt{a}$ to modality $\mathtt{b}$, indicating that the fusion of two modalities is asymmetric. Therefore, after fusing all pairs of modalities, we obtain six fused features, including $\mathbf{Z}^{RF}$, $\mathbf{Z}^{FR}$, $\mathbf{Z}^{RA}$, $\mathbf{Z}^{AR}$, $\mathbf{Z}^{FA}$, and $\mathbf{Z}^{AF}$.

\noindent\textbf{Bottleneck Token-based Weighting.}
A straightforward way to obtain the feature fused from all modalities is to concatenate the features fused from all pairs. That is, 
\begin{equation} \label{eq:fusion_prior}
\mathbf{Z} =[\mathbf{Z}^{RF}||\mathbf{Z}^{FR}|| \mathbf{Z}^{RA}||\mathbf{Z}^{AR}||\mathbf{Z}^{FA}||\mathbf{Z}^{AF}],
\end{equation}
where $\mathbf{Z} \in \mathbb{R}^{T\times (N_F\cdot D_E)}$ and $N_F$ is the number of modality pairs. However, we realize that the bottleneck token learned at the final layer, \textit{i.e.} $\tilde{\mathbf{Z}}^{bt}_{L_M}$, can be used to measure the information content transmitted from one modality to the other. Therefore, we propose to leverage the learned bottleneck tokens to weight the features fused from all pairs of modalities, highlighting more important features. The weights are obtained by feeding the tokens into a $L_W$ layer transformer and a regressor. That is, 
\begin{equation} \label{eq:w_transformer}
	\mathbf{w} =\Theta(\text{Transformer}_{(L_W\times)}(\mathbf{Z}^{bt})),
\end{equation}
where $\mathbf{Z}^{bt}$ is a stack of the final bottleneck tokens obtained from all pairwise fusions, $\text{Transformer}_{(L_W\times)}(\cdot)$ denotes a $L_W$ layer transformer, $\Theta(\cdot)$ is a regressor implemented by a three-layer Multilayer Perceptron (MLP), and $\mathbf{w} = [w_1, ..., w_{N_F}] \in \mathbb{R}^{1\times N_F}$. Then, the weighted feature is obtained by
\begin{equation}
	\hat{\mathbf{Z}}=[w_1\mathbf{Z}^{RF}||w_2\mathbf{Z}^{FR}||w_3\mathbf{Z}^{RA}||w_4\mathbf{Z}^{AR}||w_5\mathbf{Z}^{FA}||w_6\mathbf{Z}^{AF}].
\end{equation}

\subsection{Temporal Consistency Contrast}
Although a portion of audio-visual data may be asynchronous~\cite{yu2022modality}, the majority of them are intrinsically synchronized. Moreover, when modality asynchrony exists, the self-attention based fusion scheme in our MSBT is able to align two modalities implicitly. Therefore, we expect that the pairwise fused features at the same moment are similar in semantics. To achieve this, we propose a temporal consistency contrast (TCC) loss, which attracts together all fused features at the same moment and repels them from features at different time. When a video is given, this loss is formally defined by
\begin{small}
	\begin{equation} 
		\label{eq:temporal-consistency}
		\mathcal{L}_{TCC}= - \frac{1}{N_F T}\sum\limits_{t=1}^T
		\sum\limits_{\substack{\mathtt{a},\mathtt{b}\in\mathcal{M}}}
		\sum\limits_{\substack{\mathtt{c},\mathtt{d}\in\mathcal{M}\\ \mathtt{c}\neq\mathtt{a} \vee \mathtt{b}\neq\mathtt{d}}}
		log\frac{exp(\phi(\hat{\mathbf{Z}}^{\mathtt{a}\mathtt{b}}_t, \hat{\mathbf{Z}}^{\mathtt{c}\mathtt{d}}_t)/\tau)}{\sum\nolimits^{T}_{k=1}exp(\phi(\hat{\mathbf{Z}}^{\mathtt{a}\mathtt{b}}_t, \hat{\mathbf{Z}}^{\mathtt{c}\mathtt{d}}_k)/\tau)},
	\end{equation}
\end{small}
where $\phi(\cdot,\cdot)$ computes the cosine similarity, $\tau$ is a temperature hyper-parameter and the subscript $t$ or $k$ of $\hat{\mathbf{Z}}$ denotes the corresponding snippet. 

\subsection{Network Training}
Under the weakly supervised setting, each video in the training set is annotated with a binary label $y \in \{0,1\}$ to indicate whether it is a violent video or not. When the fused features of all snippets of one video are produced, we employ a global encoder that is implemented by a $L_G$ layer transformer to model the global context of all modalities, and further use a regressor to produce anomaly scores. That is, 
\begin{equation}
	\mathbf{s}=\Omega(\text{Transformer}_{(L_G\times)}(\hat{\mathbf{Z}})),
\end{equation}
where $\mathbf{s}\in\mathbb{R}^{T}$ is the anomaly scores of all snippets in the video and $\Omega(\cdot)$ is the regressor that is implemented by a three-layer MLP. We then adopt the widely used top-$K$ MIL loss~\cite{yu2022modality,sun2023long} to train the regressor and employ the temporal consistency contrast loss for feature regularization. 
To compute the MIL loss of one video, we average the top-$K$ anomaly scores, \textit{i.e.} $\bar{s}=\frac{1}{K}\sum\nolimits_{s_i\in\mathcal{T}_k(\mathbf{s})}s_i,$
here $\mathcal{T}_K(\mathbf{s})$ indicates the set of top-$K$ scores in $\mathbf{s}$. Then the MIL loss is defined by
\begin{equation} \label{eq:MIL_loss}
	\mathcal{L}_{MIL} = -y log(\bar{s})-(1-{y})log(1-\bar{s}).
\end{equation}
The entire training loss is the combination of the Top-K MIL loss $\mathcal{L}_{MIL}$ and the temporal consistency contrast loss $\mathcal{L}_{TCC}$:
\begin{equation} \label{eq:joint_loss}
	\mathcal{L} = \mathcal{L}_{MIL} + \lambda \mathcal{L}_{TCC},
\end{equation}
where $\lambda$ is a hyper-parameter for balancing two terms.

\begin{table}[t]
	\setlength{\tabcolsep}{3.9mm}{
		\caption{The AP (\%) of model variants with or without MSBT-based fusion module and TCC loss. `R', `A', and `F' denote RGB, audio, and flow modalities, respectively.}
		\begin{tabular}{cccccc}
			\toprule
			MSBT    & TCC  &   R+A   &  R+F  &  A+F   & R+A+F        \\ 
			\midrule
			\XSolidBrush   & \XSolidBrush  & 78.51 & 78.97 & 73.85 & 80.46      \\
			\Checkmark     & \XSolidBrush  & 81.63 & 79.78 & 74.63 & 81.65      \\
			\XSolidBrush   & \Checkmark    & 78.73 & 79.75 & 74.46 & 80.58      \\ 
			\Checkmark    & \Checkmark    & \textbf{82.54}   &  \textbf{80.68} & \textbf{77.47} & \textbf{84.32}        
			\\ \bottomrule
		\end{tabular}
		\label{tab:ab_components}}
	\vspace{-14pt}
\end{table}

\begin{figure*}[t]
	\centering
	\includegraphics[width=0.98\textwidth]{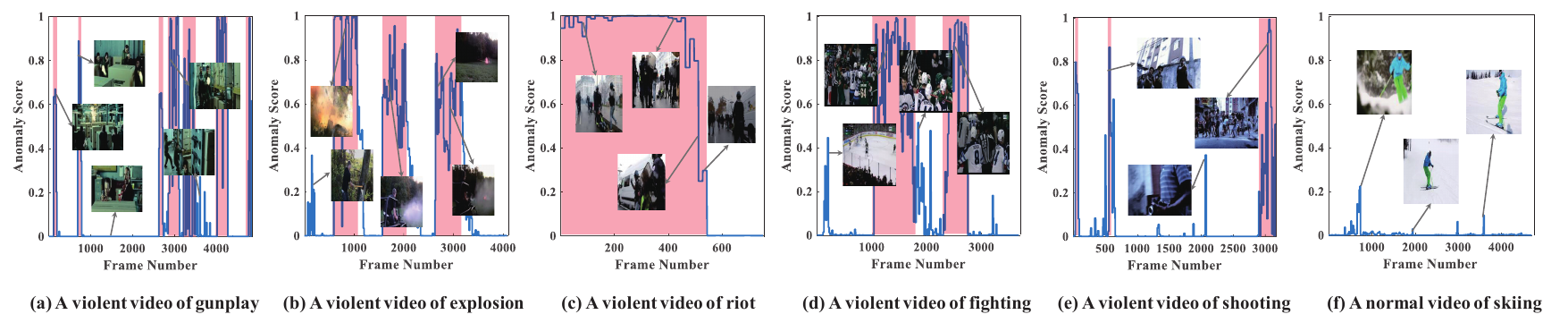}
	\caption{Visualization of anomaly scores predicted on the XD-Violence test set. The red regions indicate ground-truth violent events and the blue lines are anomaly scores predicted by our method. Best viewed in color.}
	\label{fig:anomaly_scores}
	\vspace{-14pt}
\end{figure*}

\section{Experiments}
\subsection{Dataset and Evaluation Metric}
Following state-of-the-art methods~\cite{yu2022modality,wu2020not,wu2022weakly}, we evaluate our proposed method on XD-Violence~\cite{wu2020not}, which is the largest publicly available multimodal dataset for violence detection. It consists of 4,754 untrimmed videos with a total duration of 217 hours,  covering six violence types and providing RGB, optical flow, and audio streams for each video. The dataset is collected from movies and YouTube, covering diverse scenarios and six violence types. The entire dataset is split into a training set of 3,954 videos with video-level labels and a test set of 800 videos with frame-level labels. As is common practice~\cite{yu2022modality,wu2020not,wu2022weakly}, we adopt the frame-level average precision (AP) as the evaluation metric.

\subsection{Implementation Details}
We implement our method in Pytorch. In the unimodal encoders, the dimension of encoded features is $D_E=128$, the number of attention heads is 4, and $L_U = 1$. In the MSBT, we employ $L_M = 5$ fusion layers and the number of bottleneck tokens in the first layer is $N^{bt}_1 = 16$. The learnable bottleneck tokens are initialized utilizing a Gaussian with a mean of 0 and standard deviation of 0.15, and the dimension of the bottleneck tokens is set to 128. In the module of weight computation, the number of transformer layers $L_{W}$ is set to 1. In the global encoder, we set $L_G=4$ layers for the transformer to aggregate global context. The temperature factor in the TCC loss is $\tau = 0.5$, the Top-$K$ value in the MIL loss is computed by $K = 9$, and the balancing scalar in the entire loss is $\lambda = 0.1$. Moreover, we train our model for 50 epochs, using the SGD optimizer with a batch size of 128 and a learning rate of 0.005.

\begin{table}[t]
	\setlength{\tabcolsep}{3.3mm}{
		\caption{The AP (\%) of model variants with different designs. `CT' denotes the Cross-Transformer in the MSBT, `T\_W' denotes the Transformer used in weight computation, and `Weighting' indicates that the weighting scheme is used.}
		\begin{tabular}{@{}ccccccc@{}}
			\toprule
			CT  &  T\_W  & Weighting  & R+A & R+F & A+F & R+A+F  \\ 
			\midrule
			\XSolidBrush & \XSolidBrush & \XSolidBrush & 80.86 & 79.34 & 74.15  &  80.78     \\ 
			\XSolidBrush & \XSolidBrush & \Checkmark & 81.26 & 79.63 & 74.59 & 80.83      \\ 
			\XSolidBrush & \Checkmark   & \Checkmark & 81.43 & 79.68 & 75.17 & 81.14    \\
			\midrule
			\Checkmark   & \XSolidBrush & \XSolidBrush & 81.35 & 79.55 & 74.71 & 81.32  \\
			\Checkmark   & \XSolidBrush & \Checkmark  & 81.50 & 79.82 & 74.79 & 81.34   \\  
			\Checkmark   & \Checkmark   & \Checkmark & \textbf{82.54} & \textbf{80.68} & \textbf{77.47} & \textbf{84.32}        
			\\ \bottomrule
		\end{tabular}
		\label{tab:ab_MBF}}
	\vspace{-6pt}
\end{table}

\begin{table}[t]
	\setlength{\tabcolsep}{5.4mm}{
		\caption{The AP (\%) of model variants with a fixed or reduced number of bottleneck tokens used in the MSBT. The performance is compared for the model using three modalities. 2, 4, 8, or 16 indicates the number of tokens used at the first layer (reducing case) or at all layers (fixed case).}
			\begin{tabular}{@{}ccccc@{}}
				\toprule
				Method  &  2   & 4  & 8   & 16       \\ 
				\midrule
				Fixed number &  80.61  & 81.79       & 82.12   &  82.33          \\
				Reduced number & \textbf{81.25}  & \textbf{82.46}   & \textbf{82.87}  & \textbf{84.32}   
				\\ \bottomrule
			\end{tabular}
			\label{tab:AP_fixed_multiscale}}
		\vspace{-8pt}
	\end{table}

\subsection{Ablation Studies}
To validate the effectiveness of the proposed method, we carefully examine the design of each module and investigate the performance of model variants in the following sub-sections. Moreover, since our model can deal with a varying number of modalities, we conduct experiments based on any two RGB, flow, and audio modalities, as well as the one using all three modalities.

\noindent\textbf{Effectiveness of the Proposed Modules.} 
The major contributions of the proposed method lie in the multi-scale bottleneck transformer (MSBT) based fusion module and the temporal consistency contrast (TCC) loss. Therefore, 
we first investigate the effectiveness of the designs of MSBT and TCC by leaving out either one or both of them from our framework. When the MSBT-based fusion module is removed, we alternatively use a cross-attention scheme as in~\cite{yu2022modality} for multimodal fusion. The effectiveness of these model variants is examined for different multimodal inputs, and the performance is presented in Table~\ref{tab:ab_components}. The results show that, regardless of whether two or three modalities are input, the MSBT fusion module consistently outperforms the cross-attention fusion scheme, and the TCC loss consistently improves the performance as well. Moreover, the synergy of MSBT with TCC boosts performance significantly.

\noindent\textbf{Effectiveness of the Designs in Multimodal Fusion.}
In the MSBT, our current design contains two characteristics. 1) The bottleneck tokens at each layer are passed from those learned at the previous layer via the Cross-Transformer (CT) defined in Eq.~\ref{eq:cross_transformer}. An alternative is leaving out the Cross-Transformer. That is, the tokens are randomly initialized at the beginning of each layer, and no information is passed from the previous layer. Results in Table~\ref{tab:ab_MBF} demonstrate the superiority of using CT. 2) The number of bottleneck tokens at consecutive layers of our current model is reduced by half to implement the gradual condensation. An alternative is keeping the token number fixed like~\cite{nagrani2021attention,Yan2022Transformer}. Results in Table~\ref{tab:AP_fixed_multiscale} validate the superiority of our gradual condensing scheme. In the bottleneck token-based weighting sub-module, we evaluate two alternative designs. The first design leaves out the Transformer used in Eq.~\ref{eq:w_transformer}, while the second one completely removes the weights. Results in Table~\ref{tab:ab_MBF} show that the weighting scheme helps to improve the performance.

\begin{figure}[t]
	\centering
	\includegraphics[width=0.99\linewidth]{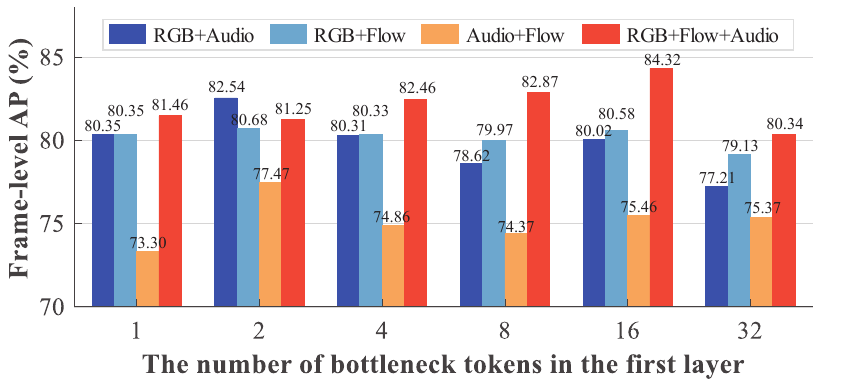}
	\caption{The performance evaluation with a different number of tokens at the first layer of our multi-scale bottleneck transformer. Best viewed in color.}
	\label{fig:bottleneck_AP}
	\vspace{-8pt}
\end{figure}

\noindent\textbf{Impact of the Bottleneck Token Number.}
The bottleneck tokens play an important role in our multimodal fusion. Therefore, we conduct experiments to check how the number of bottleneck tokens impacts the performance. To this end, 
we vary the token number $N^{bt}_1$ from 1 to 32, plotting the performance in Fig.~\ref{fig:bottleneck_AP}. The results suggest that the more modalities are used, the more bottleneck tokens are needed.

\subsection{Qualitative Analysis}
Fig.~\ref{fig:anomaly_scores} demonstrates the anomaly scores predicted by our proposed model using three modalities. It shows that our method is able to discriminate various types of violent events. Specifically, our method can identify not only short-term violent events that last for only tens of frames, such as those in Fig.~\ref{fig:anomaly_scores} (a) and (e), but also long-term violent events whose duration is hundreds of frames, like the events shown in Fig.~\ref{fig:anomaly_scores} (c) and (d). Moreover, it produces very low anomaly scores for normal events in both violent and normal videos, indicating superior discrimination ability. 

\subsection{Comparison to State-of-the-Art}
We finally compare our proposed model to state-of-the-art methods, as shown in Table~\ref{exp:table1}. Most multimodal methods focus on audio-visual learning, in which only two modalities are considered and their frameworks are not readily extendable to incorporate more modalities. When considering only two modalities, our method outperforms most counterparts, except for~\cite{yu2022modality}~(Full), which additionally trains a single-modality teacher network for knowledge distillation. The knowledge transferred from the unimodal network helps it achieve superior performance when leveraging RGB and audio modalities, but inferior performance when using other combinations of two modalities. Ultimately, our method surpasses all existing methods when all three modalities are taken into account.

\begin{table}[t]
	\setlength{\tabcolsep}{8.2mm}{
		\caption{The comparison of AP (\%) with SoTA methods. $*$ indicates that the results are obtained by re-training the model with different input modalities. The best result under the same settings is \textbf{bolded} and the second best is \underline{underlined}.}
		\begin{tabular}{@{}ccc@{}}
			\toprule
			Method     & Modality       & AP (\%)  \\ 
			\midrule
			HL-Net~\cite{wu2020not}   & RGB+Audio        & 78.64      \\
			ACF~\cite{wei2022look}  & RGB+Audio      & 80.13       \\ 
			Pang~\textit{et al.}~\cite{pang2022audiovisual} & RGB+Audio      & 79.37    \\
			Zhang~\textit{et al.}~\cite{zhang2023exploiting} & RGB+Audio    &  81.43    \\
			Yu~\textit{et al.}~\cite{yu2022modality}~(Light) & RGB+Audio      & 82.17         \\ 
			Yu~\textit{et al.}~\cite{yu2022modality}~(Full) & RGB+Audio      & \textbf{83.40}         \\
			Yu~\textit{et al.}~\cite{yu2022modality}(Light)*   & RGB+Flow      & 78.49        \\ 
			Yu~\textit{et al.}~\cite{yu2022modality}~(Full)* & RGB+Flow      &  \underline{79.73}        \\ 
			Yu~\textit{et al.}~\cite{yu2022modality}(Light)*   & Audio+Flow      & 76.58         \\
			Yu~\textit{et al.}~\cite{yu2022modality}~(Full)* & Audio+Flow      & \underline{77.06}         \\ 
			Wu~\textit{et al.}~\cite{wu2022weakly} & Audio+Flow     & 72.96         \\ 
			Wu~\textit{et al.}~\cite{wu2022weakly} & RGB+Audio+Flow     & 79.53 \\
			Xiao~\textit{et al.}~\cite{xiao2022optical} & RGB+Audio+Flow     & \underline{83.09} \\ 

			\midrule
			MSBT      & RGB+Audio            & \underline{82.54}  \\ 
			MSBT           & RGB+Flow            & \textbf{80.68}  \\
			MSBT           & Audio+Flow            & \textbf{77.47}  \\
			MSBT         & RGB+Audio+Flow            & \textbf{84.32}   \\ 
			\bottomrule
		\end{tabular}
		\label{exp:table1}}
\end{table}

\section{Conclusion}
In this work, we have presented a new method for multimodal violence detection. It constructs a fully transformer-based network architecture and adopts the MIL-based framework for weakly supervised learning. Besides, a multi-scale bottleneck transformer (MSBT) based fusion is designed to effectively fuse multiple modalities and a temporal consistency contrast (TCC) loss is devised to better learn discriminative features. Our experiments on XD-violence~\cite{wu2020not} have validated the effectiveness of the proposed method. The experimental results show that our method achieves state-of-the-art performance, especially when three modalities are all considered. Moreover, our method is also extendable to any number of input modalities, potentially allowing for the inclusion of more modalities such as depth or infrared video streams in the future.

\section*{Acknowledgment}
This research was funded by the Zhejiang Province Pioneer Research and Development Project ``Research on Multi-modal Traffic Accident Holographic Restoration and Scene Database Construction Based on Vehicle-cloud Intersection'' (Grant No. 2024C01017).

\bibliographystyle{IEEEtran}
\bibliography{ICME2024_MSBT}

\end{document}